\documentclass[conference]{IEEEtran}
\IEEEoverridecommandlockouts
\usepackage{cite}
\usepackage{amsmath,amssymb,amsfonts}
\usepackage{algorithmic}
\usepackage{graphicx}
\usepackage{textcomp}
\usepackage{xcolor}
\usepackage{cite}
\usepackage{subfig}
\usepackage{amsmath,amssymb,amsfonts}
\usepackage[ruled,linesnumbered]{algorithm2e}
\usepackage{multirow,booktabs,color,soul,threeparttable}
\usepackage{bm}

\newcommand{\RNum}[1]{\uppercase\expandafter{\romannumeral #1\relax}}
\definecolor{hl}{rgb}{0.75,0.75,0.75}
\sethlcolor{hl}

\def\BibTeX{{\rm B\kern-.05em{\sc i\kern-.025em b}\kern-.08em
    T\kern-.1667em\lower.7ex\hbox{E}\kern-.125emX}}

  \makeatletter
  \newcommand{\linebreakand}{%
    \end{@IEEEauthorhalign}
    \hfill\mbox{}\par
    \mbox{}\hfill\begin{@IEEEauthorhalign}
  }
  \makeatother

\begin{document}
\title{Model Uncertainty in Evolutionary Optimization and Bayesian Optimization: A Comparative Analysis
\thanks{This work is supported by the National Natural Science Foundation of China~(No.62306174), the China Postdoctoral Science Foundation~(No.2023M74225, No.2023TQ0213) and the Postdoctoral Fellowship Program of CPSF under Grant Number~(No.GZC20231588).}}

% \author{\IEEEauthorblockN{Anonymous Authors}}
\author{
\IEEEauthorblockN{Hao~Hao\IEEEauthorrefmark{1}, Xiaoqun~Zhang\IEEEauthorrefmark{1}, Aimin~Zhou\IEEEauthorrefmark{2}}
\IEEEauthorblockA{
  \IEEEauthorrefmark{1}\textit{Institute of Natural Sciences, Shanghai Jiao Tong University, Shanghai, China} \\
  \IEEEauthorrefmark{2}\textit{School of Computer Science and Technology, East China Normal University, Shanghai, China}\\
haohao@sjtu.edu.cn, xqzhang@sjtu.edu.cn,  amzhou@cs.ecnu.edu.cn}
}

\maketitle

\begin{abstract}
Black-box optimization problems, which are common in many real-world applications, require optimization through input-output interactions without access to internal workings. This often leads to significant computational resources being consumed for simulations. Bayesian Optimization (BO) and Surrogate-Assisted Evolutionary Algorithm (SAEA) are two widely used gradient-free optimization techniques employed to address such challenges. Both approaches follow a similar iterative procedure that relies on surrogate models to guide the search process. This paper aims to elucidate the similarities and differences in the utilization of model uncertainty between these two methods, as well as the impact of model inaccuracies on algorithmic performance. A novel model-assisted strategy is introduced, which utilizes unevaluated solutions to generate offspring, leveraging the population-based search capabilities of evolutionary algorithm to enhance the effectiveness of model-assisted optimization. Experimental results demonstrate that the proposed approach outperforms mainstream Bayesian optimization algorithms in terms of accuracy and efficiency.
  \end{abstract}

\begin{IEEEkeywords}
Black-box optimization, Surrogate-Assisted Evolutionary Algorithm, Bayesian Optimization, Model Uncertainty
\end{IEEEkeywords}

\section{Introduction}
Black-box optimization is a crucial technique used to optimize objective functions in various applications. The term ``black-box" emphasizes the lack of access to derivative information, such as gradients, which necessitates the use of algorithms that adaptively select inputs for evaluation. The goal is to efficiently search for the global optimum. This approach is essential in fields like machine learning, where fine-tuning models with opaque internal weights is required~\cite{pmlr-v176-gasse22a}, as well as in antenna design~\cite{hao2022expensive} and optimization of aerodynamic shapes in computational fluid dynamics (CFD), where the relationship between design parameters and performance is complex and highly non-linear~\cite{JinSurrogateassistedevolutionarycomputation2011,yang2024reducing}. In each case, the optimization process aims to maximize or minimize outputs based on inputs, navigating the search space with a focus on rapid convergence to the best solution while minimizing the number of evaluations required.

Bayesian Optimization (BO)~\cite{DBLP:conf/gecco/AhnRG04} and Surrogate-Assisted Evolutionary Algorithm (SAEA)~\cite{JinSurrogateassistedevolutionarycomputation2011,hao2024enhancing} both significantly enhance the efficiency of black-box optimization. BO employs probabilistic models, such as Gaussian Processes or Random Forests, to estimate the objective function and uses an acquisition function to decide where to sample next. This technique effectively balances the need for exploration against the need for exploitation. Similarly, SAEA utilize surrogate models to approximate the objective function based on previous evaluations, guiding the evolutionary search toward promising areas and reducing the need for costly function evaluations~\cite{hao2018comparison,hao2020binary}. Both strategies are designed to converge on optimal solutions with fewer evaluations, which is particularly beneficial for problems where assessing the objective function is expensive.

Surrogate models are integral to both BO and SAEA, serving as replacements for computationally expensive black-box functions. Due to the intrinsic challenge that no model can perfectly replicate the target function, it becomes essential to evaluate and manage model errors. Uncertainty quantification plays a pivotal role in this context, providing a measure of the surrogate model's error. BO and SAEA, while sharing similar search strategies, face parallel challenges. Among these are the computational burden associated with Gaussian Process (GP) modeling in medium to high-dimensional spaces, the precision of uncertainty estimation in the surrogate models, and the development of effective acquisition functions for model management~\cite{DBLP:journals/csur/WangJSO23}. Each of these issues is crucial to the performance and efficiency of the optimization algorithms, and addressing them can lead to more robust and faster convergence to the global optimum.

The research and application of model uncertainty have seen a wealth of excellent contributions, yet to this point, no studies have conducted a comparative exploration of this aspect between BO and SAEA. Moreover, an inherent distinction exists between BO and SAEA: BO is typically viewed as a single-point (sequential) search algorithm, necessitating high-quality surrogate models and precise uncertainty assessments to steer its search. Conversely, SAEA operate on a population-based stochastic search paradigm. This distinction prompts the question: Is model uncertainty still a critical component in SAEA? How may we avert an excessive dependence on model accuracy, as well as a reliance on model uncertainty, which are difficult to achieve in high-dimensional spaces? This study aims to delineate the differences between BO and SAEA, as well as to investigate the role of model uncertainty within these two algorithmic frameworks. The contributions of this work are as follows:

\begin{enumerate}
\item This investigation endeavors to bridge the research gaps within the comparative domain of Bayesian optimization and model-assisted evolutionary computation. By employing a harmonized testing framework, the study meticulously contrasts the performance nuances of quintessential algorithms from each paradigm, thereby providing a detailed exposition of their inherent proficiencies and limitations.
\item The research elucidates the significance of model uncertainty, delineating its non-essentiality in the context of evolutionary algorithm~(EA) while highlighting its criticality within Bayesian optimization. This discourse illuminates the relative merits and constraints inherent to both methodologies, offering a discerning understanding of their differential applications.
\item The work introduces a novel, generalized model management strategy specifically tailored for evolutionary computation, ingeniously conceived to circumvent the ``curse of dimensionality'' that typically plagues the computation of high-fidelity uncertainties.
\end{enumerate}

The organization of the remainder of this paper is as follows. Section~\ref{sec:preliminaries} provides the preliminaries, offering a clear introduction to the problems addressed and the algorithms involved. Section~\ref{sec:comparison} conducts a comparative analysis of BO and SAEA, highlighting their similarities and differences, and proposes a novel strategy. Section~\ref{sec:experiments} details the systematic experiments conducted to evaluate the proposed strategy. Finally, Section~\ref{sec:conclusion} summarizes the findings and discusses potential avenues for future research.

\section{Preliminaries}
\label{sec:preliminaries}
\subsection{Expensive Black-box Optimization Problem}

Consider a black-box function $f: \mathbb{R}^n \rightarrow \mathbb{R}$, where each evaluation of $f$ is assumed to be expensive in terms of computational resources or time. The objective of the black-box optimization problem is to find~\cite{DBLP:conf/kdd/GolovinSMKKS17}:

\begin{equation}
\mathbf{x}^* = \arg\min_{\mathbf{x} \in \mathcal{X}} f(\mathbf{x})
\end{equation}
where:
\begin{itemize}
  \item $\mathbf{x} \in \mathbb{R}^n$ represents a vector of decision variables.
  \item $\mathcal{X} \subseteq \mathbb{R}^n$ denotes the feasible region within the decision variable space.
  \item $f(\mathbf{x})$ is the value of the objective function for a given vector $\mathbf{x}$, which is costly to evaluate.
  \item $\mathbf{x}^*$ is the optimal solution, satisfying $f(\mathbf{x}^*) \leq f(\mathbf{x})$ for all $\mathbf{x} \in \mathcal{X}$.
\end{itemize}

The function $f$ is considered a black box as its explicit form is unknown, rendering gradient-based optimization methods inapplicable. Evaluating $f(\mathbf{x})$ typically demands substantial computational effort. The challenge lies in identifying the optimal $\mathbf{x}^*$ with the fewest possible evaluations of $f$, owing to the significant cost associated with each evaluation.

\subsection{Typical Algorithmic Frameworks}

As quintessential methodologies for solving expensive black-box optimization problems, BO and SAEA share considerable common ground in both conceptual ideology and structural framework. This section is dedicated to presenting the fundamental frameworks of these two categories of algorithms, thereby setting the stage for subsequent comparative analysis. Algorithm~\ref{alg:BO} and Algorithm~\ref{alg:SAEA} outline the respective foundational frameworks of BO and SAEA.

\subsubsection{Bayesian Optimization}

\begin{algorithm}[htbp]
  \caption{Bayesian Optimization (BO)} \label{alg:BO}
  \KwIn{Objective function $f(\mathbf{x})$}
  \KwOut{The best solution $\mathbf{x}^{*}$ and its objective value $f(\mathbf{x}^{*})$}
  Initialize observed data $\mathbf{D}$, select a surrogate model $\mathcal{M}$, and choose an acquisition function\; \label{alg:BO-init}
  \While{stopping criterion not met}{
    Fit $\mathcal{M}$ to $\mathbf{D}$\; \label{alg:BO-fit}
    Select the next query point $\mathbf{x}_{next}$ using the acquisition function\; \label{alg:BO-acq}
    Evaluate $f(\mathbf{x}_{next})$ and augment $\mathbf{D}$ with the new observation $(\mathbf{x}_{next}, f(\mathbf{x}_{next}))$\; \label{alg:BO-eval}
  }
\end{algorithm}

In Algorithm~\ref{alg:BO}, BO begins by initializing a set of observed data points, choosing a surrogate model, and selecting an acquisition function~\cite{DBLP:conf/nips/WilsonHD18}(line~\ref{alg:BO-init}). It then fits the surrogate model to the observed data (line~\ref{alg:BO-fit}). Based on the fitted model, the algorithm computes the acquisition function to identify the next sampling point $\mathbf{x}_{next}$ (line~\ref{alg:BO-acq}). Following this, the objective function $f$ is evaluated at $\mathbf{x}_{next}$, and the observation is incorporated into the dataset $\mathbf{D}$. This iterative process of updating the model and acquiring new samples continues until a predefined stopping criterion is fulfilled, such as reaching a maximum number of iterations. The algorithm concludes by returning the best solution $\mathbf{x}^{*}$ and its associated objective value $f(\mathbf{x}^{*})$, which signifies the most favorable result discovered within the explored search space.

\subsubsection{Surrogate-assisted Evolutionary Algorithm}

\begin{algorithm}[htbp]
\caption{Surrogate-assisted Evolutionary Algorithm (SAEA)} \label{alg:SAEA}
\KwIn{Objective function $f(\mathbf{x})$, population size $N$}
\KwOut{The best solution $\mathbf{x}^*$ and its fitness $f(\mathbf{x}^*)$}
Initialize a population $\mathbf{P}$ of $N$ individuals, evaluate their fitness with $f$, start surrogate model $\mathcal{M}$, and form dataset $\mathbf{D}$\; \label{alg:SAEA-init}
  \While{stopping criterion not met}{
      Train $\mathcal{M}$ using $\mathbf{D}$\; \label{alg:SAEA-fit}
      Generate offspring $\mathbf{O}$ by applying reproductive operators to $\mathbf{P}$\; \label{alg:SAEA-gen}
      Select a subset of $\mathbf{O}$ for real evaluation based on $\mathcal{M}$ predictions and evaluate their fitness with $f$\; \label{alg:SAEA-sel}
      Integrate the newly evaluated individuals into $\mathbf{P}$ and update the dataset $\mathbf{D}$\; \label{alg:SAEA-update} \label{alg:SAEA-eval}
  }
\end{algorithm}

In Algorithm~\ref{alg:SAEA}, SAEA commences with the initialization of a population $\mathbf{P}$ of $N$ individuals, assessing their fitness using the objective function $f$. The population $\mathbf{P}$ typically forms the initial training set $\mathbf{D}$ for the surrogate model $\mathcal{M}$ (line~\ref{alg:SAEA-init}). With each iteration, $\mathcal{M}$ is trained on the dataset $\mathbf{D}$ to approximate the fitness landscape (line~\ref{alg:SAEA-fit}). Offspring $\mathbf{O}$ are produced from $\mathbf{P}$ using reproductive operators such as selection, crossover, and mutation (line~\ref{alg:SAEA-gen}). A subset of these offspring is then selected for true fitness evaluation, guided by the predictions of $\mathcal{M}$ (line~\ref{alg:SAEA-sel}). The evaluated individuals are integrated into the population $\mathbf{P}$, and the dataset $\mathbf{D}$ is updated accordingly (line~\ref{alg:SAEA-update}). This process is repeated until a stopping criterion is reached, for example, a maximum number of fitness evaluations (FEs). The algorithm yields the best solution $\mathbf{x}^*$ found during the evolutionary process, along with its fitness value $f(\mathbf{x}^*)$.

Both algorithms rely on model-based optimization techniques, which require updating the model at each iteration to inform the subsequent search phase. Although there are similarities in their reliance on a model to guide the search, the specific application and integration of the model within each algorithm differ, which will be explored in detail in the following section.

\begin{figure*}[htbp]
  \centering
  \includegraphics[width=0.95\textwidth]{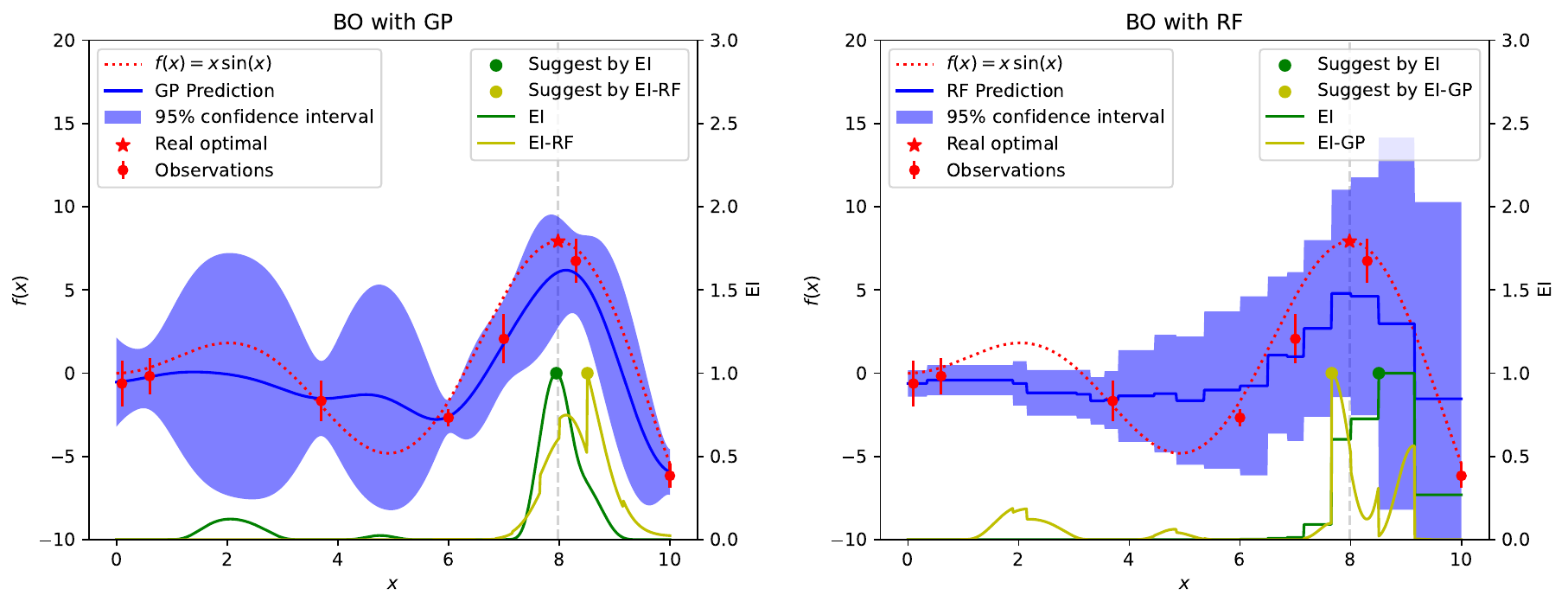}
  \caption{Performance comparison of GP and RF in uncertainty estimation and function fitting in BO.}
  \label{fig:bo2d}
\end{figure*}

\begin{figure*}[htbp]
  \centering
  \includegraphics[width=0.95\textwidth]{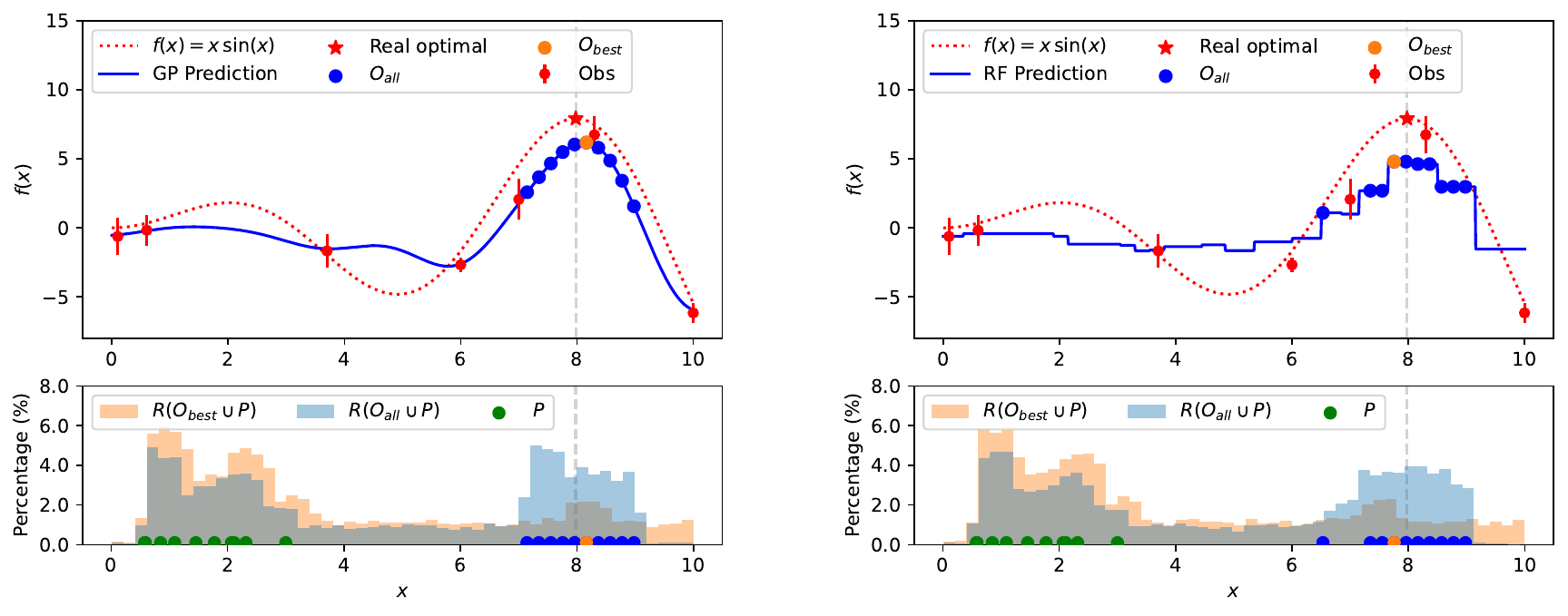}
  \caption{Performance comparison of GP and RF in function fitting and new solution reproduction in SAEA.}
  \label{fig:ea2d}
\end{figure*}

\section{Model Uncertainty in BOs and SAEAs}
\label{sec:comparison}
\subsection{Comparative Analysis}
This subsection delves into the search patterns and the impact of model uncertainty within the frameworks of BO and SAEA through low-dimensional visualization. We consider the function defined in Equation~(\ref{euq:simplefx}), which has its peak at $x = 7.99$, yielding $f(x) = 7.92$, where $x$ is in the range $[0,10]$. The surrogate models selected are Gaussian Processes (GPs)~\cite{DBLP:conf/ac/Rasmussen03} and Random Forests (RFs)~\cite{BreimanRandomForests2001}, both of which are extensively employed in the domains of BO and SAEA. Seven observational data points are initially provided, each subject to stochastic perturbations, with the random seed being held constant. The resulting visualizations are depicted in Figures~\ref{fig:bo2d} and \ref{fig:ea2d}.

\begin{equation}
f(x) = x\sin(x)
\label{euq:simplefx}
\end{equation}

\subsubsection{BO Framework}
Fig.~\ref{alg:BO} illustrates the BO process (Algorithm~\ref{alg:BO}), highlighting how a surrogate model is constructed from observational data (line~\ref{alg:BO-fit}) and how the acquisition function, specifically Expected Improvement (EI)~\cite{DBLP:conf/ifip7/Mockus74}, guides the selection of points for subsequent iterations (line~\ref{alg:BO-acq}).

Fig.~\ref{fig:bo2d} reveals that the GP outperforms the RF in terms of function fitting, providing a more accurate estimation of uncertainty. For example, in the interval from 0 to 4, there is a significant discrepancy between the fitted curve and the true function; the GP model accurately estimates this divergence, while the RF model does not. Conversely, in the interval from 6 to 10, where the divergence between the fitted and true curves is minimal, the GP model exhibits lower uncertainty, whereas the uncertainty predicted by the RF model remains relatively high. Further analysis is provided by the green curve in the diagram, which represents the values of EI; a higher value of EI indicates greater exploratory potential at that point. The next point to be sampled, $\mathbf{x}_{next}$, selected based on the GP model, is closer to the real optimum, whereas the $\mathbf{x}_{next}$ chosen using the RF model deviates from the true optimum. Additionally, by exchanging the uncertainty estimations of the two models, we obtain a comparative curve, EI*. The figure clearly shows that when the GP model is misled by inaccurate uncertainty estimation, there is a distortion in the acquisition function selection. Similarly, the RF model, hindered by its poor fitting capabilities, experiences a shift in acquisition choice, even when it employs a precise estimate of uncertainty.

Therefore, it is evident that accurate function fitting is fundamental to the operation of BO, while a reliable assessment of uncertainty is crucial for the accurate search of the acquisition function.

\subsubsection{SAEA Framework}
The essence of EA lies in the execution of selection and generation, methodologies that are fundamentally population-based. Fig.~\ref{fig:ea2d} illustrates this process. Within the figure, the blue points (denoted as $\mathbf{O}_{all}$) represent the top 20\% of points, chosen from a uniformly distributed sample of 100 points within the 0-10 interval based solely on the model's predicted values. The orange points represent the optimal points as perceived by the surrogate model. Although the fitting capabilities of the RF is inferior by those of the GP, the RF still captures the essential features of the functional landscape. Consequently, this has a minimal impact on the selection from $\mathbf{O}_{all}$, which underscores the robustness of population-based selection strategies in overcoming model inaccuracies.

Subsequently, the process of reproduction takes place. We investigate the effects that the number of offspring ($\mathbf{O}_{all}$ or $\mathbf{O}_{best}$), as selected by the surrogate model, has on the distribution of new solutions in the following iteration. The green point $\mathbf{P}$ represents the current distribution of the parental population. When combined separately with $\mathbf{O}_{best}$ and $\mathbf{O}_{all}$, and employing the reproduction operator~\cite{journals/tec/ZhouSZ15}, 10,000 new solutions were sampled, as depicted by the orange and blue shaded areas in Fig.~\ref{fig:ea2d}. The results show that new solutions derived from $\mathbf{P} \cup \mathbf{O}_{all}$ have a significantly higher probability of appearing near the optimal region compared to those from $\mathbf{P} \cup \mathbf{O}_{best}$, regardless of whether the GP or RF model is used for selection.

This indicates that updating the population with $\mathbf{P} \cup \mathbf{O}_{all}$ can effectively improve the quality of new solutions emerging from the population, while relying solely on $\mathbf{P} \cup \mathbf{O}_{best}$ may lead to a distribution of new solutions that diverges from the optimal area, thereby reducing the quality of the solutions. Furthermore, during this process, model uncertainty does not seem to significantly influence population iteration, and the difference in modeling accuracy between GP and RF appears to be mitigated, to some extent, by the evolutionary process.

\subsection{New Model Management Strategy}
In EAs, the use of a population-based search and generation methodology can mitigate the impact of model accuracy issues, necessitating reliance on a broader population ($\mathbf{O}_{all}$) rather than on isolated solutions ($\mathbf{O}_{best}$). Reproduction does not depend on the objective function $f$, making it a prominent strategy to integrate the subset $\mathbf{O}_{all}$, selected by the model, into the population without evaluated. This approach is preferred over exclusively incorporating the optimal individuals $\mathbf{O}_{best}$. This strategy not only ensures the diversity of the population but also maintains the quality of the population and the continuous updating of the model training data.

As a straightforward and universally applicable strategy, it requires only minor modifications to Algorithm~\ref{alg:SAEA}:
\begin{itemize}
  \item Line~\ref{alg:SAEA-sel}: Select a subset $\mathbf{O}_{all}$ from $\mathbf{O}$ using the surrogate model $\mathcal{M}$, and identify the most promising candidate $\mathbf{O}_{best}$ within this subset for fitness evaluation.

  \item Line~\ref{alg:SAEA-eval}: Use $\mathbf{O}_{best}$ to enhance the dataset $\mathbf{D}$, and select the top $N$ solutions from $\mathbf{D}$ to merge with $\mathbf{O}_{\text{all}}$ to form the new population $\mathbf{P}$.
\end{itemize}

In the execution of this strategy, a distribution estimation algorithm known as the variable histogram serves as the reproduction operator, a distribution estimation algorithm (EDA) known as variable-width histogram~(VWH)~\cite{journals/tec/ZhouSZ15} is utilized as the reproduction operator. Due to the inclusion of solutions that have not been evaluated, this approach is termed the unevaluated solution-derived EDA (UEDA). By default, the population size $N$ is set to 50, and $\mathbf{O}_{\text{All}}$ is set to $N/2$. The dataset $\mathbf{D}$ is pruned to maintain a maximum size of 100 through a survival-of-the-fittest selection process based on $f$.

\section{Experimental Results}
\label{sec:experiments}

\subsection{Experimental Setup}
\subsubsection{\textbf{Comparison Algorithms}}
For comparative analysis, this section identifies three representative Bayesian optimization toolkits: the BayesianOptimization toolkit (BOT)~\cite{BO2014}, Heteroscedastic Evolutionary Bayesian Optimisation (HEBO)~\cite{Cowen-Rivers2022-HEBO}, and Scikit-Optimize (Skopt)~\cite{scikit-optimize}. These toolkits are based on Bayesian optimization algorithms and possess the following principal attributes:
\begin{itemize}
  \item BOT: Incorporates sequential domain reduction~\cite{Standerrobustnesssimpledomain2002a} within the standard BO framework, significantly accelerating the search progress and hastening convergence.
  \item HEBO: Implements input and output warping to reduce the effects of heteroscedasticity and non-stationarity in the objective function. It employs multi-objective optimization to optimize multiple acquisition functions concurrently, enhancing the algorithm's robustness.
  \item Skopt: Uses the Hedge strategy~\cite{brochu2011portfolio} to probabilistically select among different acquisition functions at each iteration, increasing the robustness of the search process.
\end{itemize}

Additionally, two surrogate-assisted evolutionary algorithms (SAEAs) are utilized in the experiments: the surrogate-assisted multiswarm optimization (SAMSO)~\cite{journals/tcyb/LiCGS21} and the fuzzy classification-based preselection for CoDE (FCPS-CoDE)~\cite{conf/aaai/ZhouZSZ19}, which exemplify EAs assisted by regression and classification models, respectively. An EDA with a local search (EDA/LS)~\cite{journals/tec/ZhouSZ15} serves as the baseline algorithm for this study and is included in the experimental analysis. A detailed description is as follows:
\begin{itemize}
  \item SAMSO: Utilizes a dual-swarm approach with cross-swarm intelligence sharing and dynamic swarm size regulation, guided by a Radial Basis Function (RBF) to estimate the evaluation function.
  \item FCPS-CoDE: Employs a fuzzy K-nearest neighbors (KNN) classifier to predict the solution categories before evaluation, aiding the Composite Differential Evolution (CoDE) algorithm in its evolutionary search.
  \item EDA/LS: Integrates an Estimation of Distribution Algorithm (EDA) using a Variable Width Histogram (VWH) and includes a local search operator to facilitate the evolutionary search.
\end{itemize}

The innovations of this work, which incorporate Random Forest (UEDA-RF), Extreme Gradient Boosting~\cite{DBLP:conf/kdd/ChenG16}(UEDA-XG), and Gaussian Process (UEDA-GP), also participate in the experiments.

\subsubsection{\textbf{Test Suites}}
The test suites include the LZG~\cite{journals/tec/LiuZG14} and YLL~\cite{yao1999evolutionary} test sets. LZG comprises functions like Ellipsoid, Rosenbrock, Ackley, and Griewank, which present diverse landscapes, including unimodal, gully, and multimodal terrains. YLLF01-13 features characteristics such as unimodal, multimodal, stepwise, and stochastic noise (excluding YLLF10-11 due to overlap with LZG).

\subsubsection{\textbf{Parameter Settings}}
To ensure fair comparison, the recommended parameters from the original literature are adopted for each algorithm~\footnote{The Bayesian optimization packages are used with default settings; SAMSO is implemented within Platemo~\cite{tian2017platemo}; FCPS-CoDE and EDA/LS are realized by us, following the original documentation.}. The specific parameters are as follows:
\begin{itemize}
  \item Maximum evaluation budget: 500.
  \item Population size: Set to $N=30$ for EDA/LS and FCPS-CoDE, $N=40$ for SAMSO, and $N=50$ for UEDA-GP, UEDA-XG, and UEDA-RF.
  \item Dimensionality: All test problems are set with dimensions $n=20$ and $n=50$.
  \item Surrogate model: GPs are used for $n=20$ in the three BO toolkits. For $n=50$, RFs are preferred due to lower computational costs. (BOT does not offer an RF option.)
\end{itemize}

Each algorithm undergoes 30 independent runs on each test instance to accommodate randomness. The Wilcoxon rank-sum test~\cite{hollander2013nonparametric} is applied to analyze the results. In the tables, the symbols `+', `-', and `$\thicksim$' indicate comparative performance against the benchmark algorithms (UEDA-GP for $n=20$ and UEDA-RF for $n=50$), with `+' denoting significantly better performance, `-' indicating significantly worse, and `$\thicksim$' signifying no significant difference at the 0.05 significance level.

\subsection{Performance Analysis}

This section compares the performance of various algorithms across two aspects: solution accuracy and runtime. Table~\ref{tab:comparison} displays the problem-solving precision of each algorithm at varying dimensions. Observations from the table suggest that for lower-dimensional problems (n=20), GPs and XGBoost-assisted UEDAs surpass the RF model in effectiveness but fall short when compared to HEBO-GP. Among nine algorithms, HEBO-GP achieved an mean rank of 1.73 across 15 test problems, demonstrating a distinct advantage over the algorithm proposed in this chapter, UEDA-GP, which resulted in 12 superior outcomes, 2 inferior, and one approximate to HEBO-GP. UEDA-GP, while behind HEBO-GP, nonetheless showed significant superiority over other BO algorithms and also outperformed the compared EA methods. Notably, UEDA-RF has proven its prowess by outperforming BOT-GP, thereby confirming that a solution without evaluation can compensate for RF's lack of modeling accuracy, and in some instances, reach the level of BO. In the case of higher-dimensional problems (n=50), UEDA-RF achieved an mean rank of 2.8, which, from the viewpoint of statistical significance, indicates a clear advantage. Specifically, RF-UEDA was superior to GP-UEDA within the same framework and outperformed all BOs tested (even BOT using a GP) and was markedly better than EAs. Alternatively, HEBO when switching to an RF model, experienced a significant decline in performance, attaining the worst mean rank of 8.53. This suggests that the unreliable uncertainty estimates of RF models can mislead the exploratory direction of the algorithm.

This section compares the performance of various algorithms in terms of solution accuracy and runtime. Table~\ref{tab:comparison} presents the problem-solving precision of each algorithm at different dimensions. Observations from the table indicate that for lower-dimensional problems ($n=20$), GP and XGBoost-assisted UEDAs outperform the RF in effectiveness but do not match the performance of HEBO-GP. Among nine algorithms, HEBO-GP achieved a mean rank of 1.73 across 15 test problems. UEDA-GP resulted in 12 superior outcomes, 2 inferior, and one similar to HEBO-GP. While UEDA-GP trails behind HEBO-GP, it still shows significant superiority over other BOs and also outperforms the compared EAs. Notably, UEDA-RF has proven its effectiveness by outperforming BOT-GP, confirming that solutions without evaluation can compensate for RF's lack of modeling accuracy and, in some cases, achieve the level of BO. For higher-dimensional problems ($n=50$), UEDA-RF achieved a mean rank of 2.8, which statistically signifies a clear advantage. Specifically, RF-UEDA was superior within the same framework to GP-UEDA and outperformed all tested BOs  (including BOT using a GP) and was markedly better than EAs. Conversely, when HEBO switched to an RF model, there was a significant decline in performance, with the worst mean rank of 8.53. This indicates that the unreliable uncertainty estimates of RF models can mislead the exploratory direction of the algorithm.

\begin{table*}[htbp]
  \renewcommand{\arraystretch}{1.05}
  \renewcommand{\tabcolsep}{3pt}
  \centering 
  \caption{Statistics of mean and standard deviation results obtained by nine comparison algorithms on LZG and YLL test suites with $n=20,50$, adhere to a maximum evaluation budget of 500.} \scriptsize
  \begin{tabular}{c|ccc|ccc|ccc}
  \toprule
  \multicolumn{10}{c}{$n=20$} \\
  \midrule
  Problem & UEDA-GP & UEDA-XG & UEDA-RF & HEBO-GP & Skopt-GP & BOT-GP & EDA/LS & SAMSO & FCPS \\
  \hline
  \multirow{2}{*}{Ellipsoid} & 8.63e-03 [1] & 8.13e+00[5]($-$) & 1.18e+01[6]($-$) & 1.41e-01[3]($-$) & 6.58e-02[2]($-$) & 1.55e-01[4]($-$) & 7.17e+01[8]($-$) & 1.87e+01[7]($-$) & 1.30e+02[9]($-$) \\ 
    & (8.36e-03) & (4.14e+00) & (5.22e+00) & (3.84e-02) & (1.75e-02) & (2.83e-02) & (1.56e+01) & (2.52e+01) & (3.13e+01) \\  \hline
  \multirow{2}{*}{Rosenbrock} & 4.79e+01 [3] & 9.78e+01[6]($-$) & 9.68e+01[5]($-$) & 2.18e+01[1]($+$) & 5.42e+01[4]($\approx$) & 1.25e+02[7]($-$) & 2.37e+02[8]($-$) & 3.57e+01[2]($\approx$) & 3.22e+02[9]($-$) \\ 
    & (2.55e+01) & (3.99e+01) & (3.63e+01) & (1.10e+01) & (1.25e+01) & (4.24e+01) & (4.02e+01) & (2.47e+01) & (1.05e+02) \\  \hline
  \multirow{2}{*}{Ackley} & 5.29e+00 [3] & 8.17e+00[5]($-$) & 8.69e+00[6]($-$) & 9.10e-01[1]($+$) & 7.12e+00[4]($-$) & 5.19e+00[2]($\approx$) & 1.33e+01[7]($-$) & 1.83e+01[9]($-$) & 1.48e+01[8]($-$) \\ 
    & (1.37e+00) & (1.17e+00) & (1.11e+00) & (3.43e-01) & (4.46e-01) & (2.39e+00) & (7.37e-01) & (1.33e+00) & (1.00e+00) \\  \hline
  \multirow{2}{*}{Griewank} & 2.12e+01 [7] & 4.05e+00[4]($+$) & 5.33e+00[5]($+$) & 7.82e-01[1]($+$) & 1.02e+00[2]($+$) & 1.43e+00[3]($+$) & 2.96e+01[8]($-$) & 2.06e+01[6]($\approx$) & 5.46e+01[9]($-$) \\ 
    & (5.28e+00) & (1.39e+00) & (1.62e+00) & (9.12e-02) & (2.18e-02) & (1.46e-01) & (7.62e+00) & (1.33e+01) & (1.33e+01) \\  \hline
  \multirow{2}{*}{YLLF01} & 2.08e+03 [7] & 3.83e+02[4]($+$) & 4.09e+02[5]($+$) & 5.30e+00[2]($+$) & 3.66e+00[1]($+$) & 1.01e+01[3]($+$) & 3.17e+03[8]($-$) & 6.73e+02[6]($+$) & 5.37e+03[9]($-$) \\ 
    & (5.25e+02) & (2.31e+02) & (1.96e+02) & (1.70e+00) & (1.16e+00) & (2.70e+00) & (5.92e+02) & (6.43e+02) & (1.86e+03) \\  \hline
  \multirow{2}{*}{YLLF02} & 1.79e+00 [1] & 6.07e+00[3]($-$) & 3.89e+00[2]($-$) & 2.04e+01[4]($-$) & 4.62e+05[9]($-$) & 3.89e+04[8]($-$) & 2.67e+01[6]($-$) & 3.27e+01[7]($-$) & 2.41e+01[5]($-$) \\ 
    & (9.15e-01) & (1.15e+00) & (9.78e-01) & (6.62e+00) & (9.03e+05) & (1.21e+05) & (3.93e+00) & (1.72e+01) & (3.67e+00) \\  \hline
  \multirow{2}{*}{YLLF03} & 1.76e+04 [7] & 1.42e+04[4]($+$) & 8.81e+03[2]($+$) & 1.68e+04[5]($\approx$) & 7.47e+03[1]($+$) & 3.50e+04[9]($-$) & 2.32e+04[8]($-$) & 1.71e+04[6]($\approx$) & 1.26e+04[3]($+$) \\ 
    & (3.51e+03) & (3.54e+03) & (3.59e+03) & (4.33e+03) & (2.69e+03) & (6.98e+03) & (4.21e+03) & (1.38e+04) & (3.48e+03) \\  \hline
  \multirow{2}{*}{YLLF04} & 2.74e+01 [2] & 2.96e+01[5]($\approx$) & 2.97e+01[6]($\approx$) & 2.14e+01[1]($+$) & 2.78e+01[3]($-$) & 2.82e+01[4]($\approx$) & 3.28e+01[7]($-$) & 7.06e+01[9]($-$) & 4.17e+01[8]($-$) \\ 
    & (3.22e+00) & (7.88e+00) & (5.41e+00) & (1.14e+01) & (1.72e+01) & (9.43e+00) & (3.26e+00) & (7.40e+00) & (6.55e+00) \\  \hline
  \multirow{2}{*}{YLLF05} & 3.82e+04 [2] & 4.63e+04[3]($\approx$) & 8.77e+04[4]($-$) & 3.90e+02[1]($+$) & 1.77e+05[6]($-$) & 2.34e+06[8]($-$) & 1.13e+06[7]($-$) & 1.12e+05[5]($-$) & 4.71e+06[9]($-$) \\ 
    & (3.34e+04) & (3.51e+04) & (9.39e+04) & (6.20e+02) & (8.96e+04) & (1.04e+06) & (5.76e+05) & (8.27e+04) & (2.84e+06) \\  \hline
  \multirow{2}{*}{YLLF06} & 2.08e+03 [7] & 3.32e+02[4]($+$) & 5.12e+02[5]($+$) & 6.97e+00[2]($+$) & 5.30e+00[1]($+$) & 9.80e+00[3]($+$) & 3.20e+03[8]($-$) & 8.59e+02[6]($+$) & 6.26e+03[9]($-$) \\ 
    & (5.49e+02) & (1.57e+02) & (2.19e+02) & (2.09e+00) & (1.53e+00) & (3.19e+00) & (5.04e+02) & (1.24e+03) & (1.28e+03) \\  \hline
  \multirow{2}{*}{YLLF07} & 1.39e-01 [2]  & 2.43e-01[3]($-$) & 2.66e-01[4]($-$) & 9.19e-02[1]($+$) & 2.67e-01[5]($-$) & 1.05e+00[8]($-$) & 6.67e-01[7]($-$) & 3.23e-01[6]($-$) & 2.09e+00[9]($-$) \\ 
    & (5.67e-02) & (1.13e-01) & (1.57e-01) & (3.87e-02) & (9.91e-02) & (4.24e-01) & (2.75e-01) & (1.56e-01) & (9.17e-01) \\  \hline
  \multirow{2}{*}{YLLF08} & 4.97e+03 [5]  & 3.47e+03[3]($+$) & 2.49e+03[2]($+$) & 1.86e+03[1]($+$) & 5.17e+03[6]($\approx$) & 5.54e+03[8]($-$) & 4.66e+03[4]($+$) & 5.67e+03[9]($-$) & 5.52e+03[7]($-$) \\ 
    & (3.75e+02) & (5.53e+02) & (5.24e+02) & (3.48e+02) & (3.29e+02) & (5.52e+02) & (3.32e+02) & (2.83e+02) & (2.88e+02) \\  \hline
  \multirow{2}{*}{YLLF09} & 1.12e+02 [3] & 1.32e+02[6]($-$) & 1.28e+02[5]($-$) & 8.66e+01[1]($+$) & 1.63e+02[8]($-$) & 1.65e+02[9]($-$) & 1.57e+02[7]($-$) & 1.16e+02[4]($\approx$) & 1.02e+02[2]($+$) \\ 
    & (1.63e+01) & (1.55e+01) & (1.60e+01) & (1.96e+01) & (1.86e+01) & (3.53e+01) & (1.31e+01) & (5.14e+01) & (2.09e+01) \\  \hline
  \multirow{2}{*}{YLLF12} & 5.21e+03 [5] & 4.28e+02[3]($+$) & 1.87e+03[4]($+$) & 2.47e+00[1]($+$) & 6.68e+04[7]($-$) & 8.77e+06[9]($-$) & 6.09e+04[6]($-$) & 1.23e+02[2]($+$) & 2.46e+06[8]($-$) \\ 
    & (8.76e+03) & (1.76e+03) & (9.07e+03) & (1.61e+00) & (1.21e+05) & (5.76e+06) & (8.76e+04) & (5.49e+02) & (3.04e+06) \\  \hline
  \multirow{2}{*}{YLLF13} & 1.09e+11 [8]  & 6.96e+09[5]($+$) & 1.24e+10[6]($+$) & 3.17e+04[1]($+$) & 2.72e+10[7]($+$) & 8.77e+11[9]($-$) & 1.10e+06[3]($+$) & 6.82e+04[2]($+$) & 1.08e+07[4]($+$) \\ 
    & (4.44e+10) & (8.11e+09) & (1.32e+10) & (6.92e+04) & (1.23e+10) & (2.20e+11) & (5.84e+05) & (1.94e+05) & (8.07e+06) \\  \hline
  mean rank & 4.20 & 4.20 & 4.47 & 1.73 & 4.40 & 6.27 & 6.80 & 5.73 & 7.20 \\ 
  $+$ / $-$ / $\approx$ & / & 7/6/2 & 7/7/1 & 12/2/1 & 5/8/2 & 3/10/2 & 2/13/0 & 4/7/4 & 3/12/0 \\
  \midrule
  \multicolumn{10}{c}{$n=50$} \\
  \midrule
  Problem & UEDA-RF & UEDA-XG & UEDA-GP & HEBO-RF & Skopt-RF & BOT-GP & EDA/LS & SAMSO & FCPS \\
  \hline
  \multirow{2}{*}{Ellipsoid} & 8.48e+02 [3] & 1.08e+03[4]($-$) & 2.81e+02[2]($+$) & 3.59e+03[8]($-$) & 5.87e+03[9]($-$) & 7.52e+00[1]($+$) & 1.52e+03[6]($-$) & 1.31e+03[5]($\approx$) & 1.61e+03[7]($-$) \\ 
    & (1.63e+02) & (3.13e+02) & (6.92e+01) & (4.78e+02) & (4.37e+02) & (3.19e+00) & (2.23e+02) & (1.13e+03) & (3.39e+02) \\  \hline
  \multirow{2}{*}{Rosenbrock} &1.06e+03 [3] & 1.41e+03[4]($-$) & 7.00e+02[2]($+$) & 5.26e+03[8]($-$) & 9.59e+03[9]($-$) & 5.02e+02[1]($+$) & 1.78e+03[6]($-$) & 1.58e+03[5]($\approx$) & 1.90e+03[7]($-$) \\ 
    & (2.00e+02) & (5.21e+02) & (1.21e+02) & (7.33e+02) & (1.03e+03) & (1.10e+02) & (2.84e+02) & (1.70e+03) & (5.11e+02) \\  \hline
  \multirow{2}{*}{Ackley} & 1.65e+01 [3]  & 1.69e+01[4]($\approx$) & 1.63e+01[2]($+$) & 2.00e+01[8]($-$) & 2.06e+01[9]($-$) & 8.36e+00[1]($+$) & 1.76e+01[6]($-$) & 1.86e+01[7]($-$) & 1.75e+01[5]($-$) \\ 
    & (5.42e-01) & (8.67e-01) & (4.27e-01) & (2.71e-01) & (1.40e-01) & (6.22e-01) & (4.24e-01) & (1.18e+00) & (6.00e-01) \\  \hline
  \multirow{2}{*}{Griewank} & 1.52e+02 [2] & 1.82e+02[4]($-$) & 1.72e+02[3]($-$) & 5.52e+02[7]($-$) & 8.83e+02[9]($-$) & 3.74e+00[1]($+$) & 2.41e+02[5]($-$) & 6.19e+02[8]($-$) & 2.69e+02[6]($-$) \\ 
    & (2.58e+01) & (5.42e+01) & (2.49e+01) & (7.86e+01) & (5.97e+01) & (5.77e-01) & (2.86e+01) & (3.66e+02) & (6.72e+01) \\  \hline
  \multirow{2}{*}{YLLF01} & 1.53e+04 [2] & 1.99e+04[4]($-$) & 1.94e+04[3]($-$) & 6.17e+04[7]($-$) & 9.56e+04[9]($-$) & 6.58e+01[1]($+$) & 2.77e+04[5]($-$) & 6.69e+04[8]($-$) & 2.95e+04[6]($-$) \\ 
    & (2.38e+03) & (5.57e+03) & (1.99e+03) & (9.04e+03) & (8.96e+03) & (8.57e+00) & (3.92e+03) & (3.70e+04) & (5.86e+03) \\  \hline
  \multirow{2}{*}{YLLF02} & 4.81e+01 [1] & 8.01e+01[3]($-$) & 5.73e+01[2]($-$) & 3.31e+04[6]($-$) & 8.70e+17[8]($-$) & 3.06e+17[7]($-$) & 2.79e+04[5]($-$) & 1.05e+18[9]($-$) & 8.92e+01[4]($-$) \\ 
    & (4.05e+00) & (8.83e+00) & (1.15e+01) & (1.71e+05) & (4.40e+18) & (1.18e+18) & (9.04e+04) & (3.77e+18) & (8.24e+00) \\  \hline
  \multirow{2}{*}{YLLF03} & 1.06e+05 [2] & 1.32e+05[3]($-$) & 1.35e+05[5]($-$) & 1.33e+05[4]($-$) & 1.83e+05[7]($-$) & 1.85e+05[8]($-$) & 1.58e+05[6]($-$) & 2.97e+05[9]($-$) & 8.24e+04[1]($+$) \\ 
    & (1.86e+04) & (2.88e+04) & (3.00e+04) & (2.57e+04) & (3.44e+04) & (3.21e+04) & (2.51e+04) & (9.31e+04) & (1.68e+04) \\  \hline
  \multirow{2}{*}{YLLF04} & 6.15e+01 [4] & 6.17e+01[5]($\approx$) & 5.42e+01[1]($+$) & 8.08e+01[7]($-$) & 9.05e+01[9]($-$) & 7.28e+01[6]($-$) & 5.80e+01[2]($+$) & 8.74e+01[8]($-$) & 5.88e+01[3]($\approx$) \\ 
    & (3.24e+00) & (6.71e+00) & (3.58e+00) & (4.20e+00) & (2.23e+00) & (1.52e+01) & (2.78e+00) & (8.06e+00) & (5.58e+00) \\  \hline
  \multirow{2}{*}{YLLF05} & 1.40e+07 [2] & 2.36e+07[4]($-$) & 1.91e+07[3]($-$) & 1.74e+08[8]($-$) & 3.36e+08[9]($-$) & 8.64e+06[1]($+$) & 2.93e+07[5]($-$) & 1.55e+08[7]($-$) & 4.49e+07[6]($-$) \\ 
    & (5.12e+06) & (1.50e+07) & (4.17e+06) & (4.91e+07) & (3.76e+07) & (3.24e+06) & (5.30e+06) & (8.04e+07) & (1.82e+07) \\  \hline
  \multirow{2}{*}{YLLF06} & 1.67e+04 [2] & 2.06e+04[4]($\approx$) & 1.96e+04[3]($-$) & 6.22e+04[7]($-$) & 9.65e+04[9]($-$) & 7.19e+01[1]($+$) & 2.74e+04[5]($-$) & 7.82e+04[8]($-$) & 2.92e+04[6]($-$) \\ 
    & (3.02e+03) & (7.01e+03) & (2.74e+03) & (7.44e+03) & (7.35e+03) & (1.43e+01) & (4.41e+03) & (3.06e+04) & (7.35e+03) \\  \hline
  \multirow{2}{*}{YLLF07} & 9.11e+00 [3] & 1.56e+01[4]($-$) & 5.60e+00[1]($+$) & 1.25e+02[7]($-$) & 2.56e+02[9]($-$) & 8.35e+00[2]($\approx$) & 2.17e+01[5]($-$) & 1.31e+02[8]($-$) & 2.96e+01[6]($-$) \\ 
    & (2.64e+00) & (7.06e+00) & (2.34e+00) & (3.54e+01) & (3.52e+01) & (2.96e+00) & (4.75e+00) & (7.91e+01) & (1.28e+01) \\  \hline
  \multirow{2}{*}{YLLF08} & 1.39e+04 [2] & 1.51e+04[4]($-$) & 1.59e+04[6]($-$) & 1.09e+04[1]($+$) & 1.58e+04[5]($-$) & 1.69e+04[9]($-$) & 1.51e+04[3]($-$) & 1.66e+04[8]($-$) & 1.63e+04[7]($-$) \\ 
    & (6.52e+02) & (6.17e+02) & (5.00e+02) & (8.40e+02) & (6.43e+02) & (5.53e+02) & (5.84e+02) & (6.62e+02) & (6.11e+02) \\  \hline
  \multirow{2}{*}{YLLF09} & 5.24e+02 [7]  & 5.15e+02[4]($\approx$) & 4.42e+02[2]($+$) & 5.28e+02[8]($\approx$) & 6.98e+02[9]($-$) & 5.18e+02[5]($\approx$) & 5.19e+02[6]($\approx$) & 4.97e+02[3]($\approx$) & 3.80e+02[1]($+$) \\ 
    & (2.52e+01) & (3.91e+01) & (2.42e+01) & (3.68e+01) & (3.04e+01) & (3.64e+01) & (1.78e+01) & (1.11e+02) & (3.15e+01) \\  \hline
  \multirow{2}{*}{YLLF12} & 7.18e+06 [1] & 1.87e+07[4]($\approx$) & 8.64e+06[2]($-$) & 3.20e+08[7]($-$) & 7.46e+08[9]($-$) & 2.10e+07[5]($-$) & 1.64e+07[3]($-$) & 5.10e+08[8]($-$) & 5.44e+07[6]($-$) \\ 
    & (5.94e+06) & (1.88e+07) & (4.03e+06) & (1.17e+08) & (1.24e+08) & (1.73e+07) & (5.98e+06) & (3.16e+08) & (3.49e+07) \\  \hline
  \multirow{2}{*}{YLLF13} & 2.03e+12 [5] & 3.16e+12[7]($\approx$) & 2.75e+12[6]($-$) & 2.81e+13[8]($-$) & 5.42e+13[9]($-$) & 1.39e+12[4]($+$) & 7.53e+07[1]($+$) & 8.32e+08[3]($+$) & 1.58e+08[2]($+$) \\ 
    & (6.41e+11) & (1.83e+12) & (5.74e+11) & (7.18e+12) & (6.88e+12) & (6.90e+11) & (2.19e+07) & (6.39e+08) & (1.01e+08) \\  \hline
  mean rank & 2.80 & 4.13 & 2.87 & 6.73 & 8.53 & 3.53 & 4.60 & 6.93 & 4.87 \\ 
  $+$ / $-$ / $\approx$ & / & 0/9/6 & 6/9/0 & 1/13/1 & 0/15/0 & 8/5/2 & 2/12/1 & 1/11/3 & 3/11/1 \\ 
  \bottomrule
  \end{tabular}
  \label{tab:comparison}
  \end{table*}

\begin{figure}[htbp]
  \centering
  \includegraphics[width=0.45\textwidth]{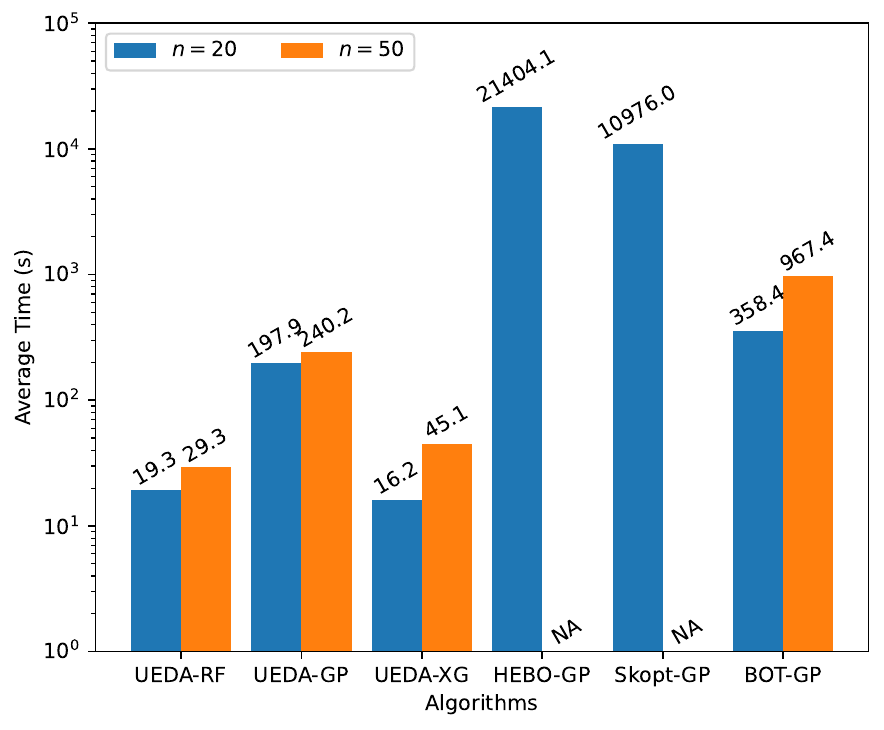}
  \caption{Runtime comparison of three BOs and UEDA with different surrogates on LZG test suits with $n=20,50$ (NA signifies an impractical computational cost for experiment completion).}
  \label{fig:run_times}
\end{figure}

Fig.~\ref{fig:run_times} presents the runtime statistics for three BOs and the UEDAs incorporating different surrogate models. The evaluations were uniformly conducted on the same hardware configuration (CPU: i9-13900k, Memory: 64GB, Ubuntu 22.04), addressing LZG01-04 problems at both 20 and 50 dimensions. This approach allowed for the determination of the average resolution times for each algorithm at different problem scales. The absence of data for HEBO-GP and Skopt-GP at $n=50$ signifies the computational cost was infeasible, hence it is denoted as `NA'.

The statistical results reveal that the UEDA algorithm significantly reduces runtime compared to the BO algorithms, particularly in comparison with HEBO-GP and Skopt-GP, by more than two orders of magnitude. This efficiency is attributed to UEDA's use of a more lightweight surrogate model and a reduced amount of model training data.

The experiments lead to the conclusion that while BO algorithms maintain a significant edge in low-dimensional problems, the modeling process involving GPs requires a substantial computational budget. In higher-dimensional, GPs are less applicable due to the curse of dimensionality, while RFs can somewhat mitigate running costs. However, RFs' imprecision in fitting and uncertainty assessment may mislead the optimization process, leading to a significant decline in BO performance. Conversely, the UEDA framework proposed in this study exploits the characteristics of population-based searches by incorporating information from unevaluated solutions to enhance search efficiency. It also leverages the strengths of RFs to reduce running costs and compensates for issues with modeling accuracy, while avoiding reliance on inaccurate uncertainty assessments, thus exhibiting robust advantages across low to moderate dimensional spaces.

\subsection{Ablation Study}

This section substantiates the effectiveness of the presented UEDA framework~(utilizing RF as the surrogate model) through ablation studies. Variants are described as presented in Table~\ref{tab:ablation}. The ablation experiments investigate the advantages of UEDA in comparison to the baseline EDA/LS, as well as the impact on the algorithm's performance of evaluating all solutions selected by the surrogate model and the exclusion of unevaluated solutions in reproduction. The experiments are conducted on the LZG test suite, with problem dimension settings of $n=20,50$. Each experimental set is independently replicated 30 times to mitigate randomness. 

\begin{table}[ht!]
  \centering
  \caption{Algorithm variants for ablation study.}
  \begin{tabular}{ll}
  \toprule
  Algorithm & Description \\
  \midrule
  UEDA-RF    & UEDA with RF surrogate \\
  UEDA-RF-AL & Evaluate all solutions selected by the model \\
  UEDA-RF-NS & Excludes unevaluated solutions from reproduction \\
  EDA/LS~\cite{journals/tec/ZhouSZ15}     & The baseline algorithm for UEDA \\
  \bottomrule
  \end{tabular}
  \label{tab:ablation}
  \end{table}

  \begin{figure}[ht!]
    \centering
    \includegraphics[width=0.5\textwidth]{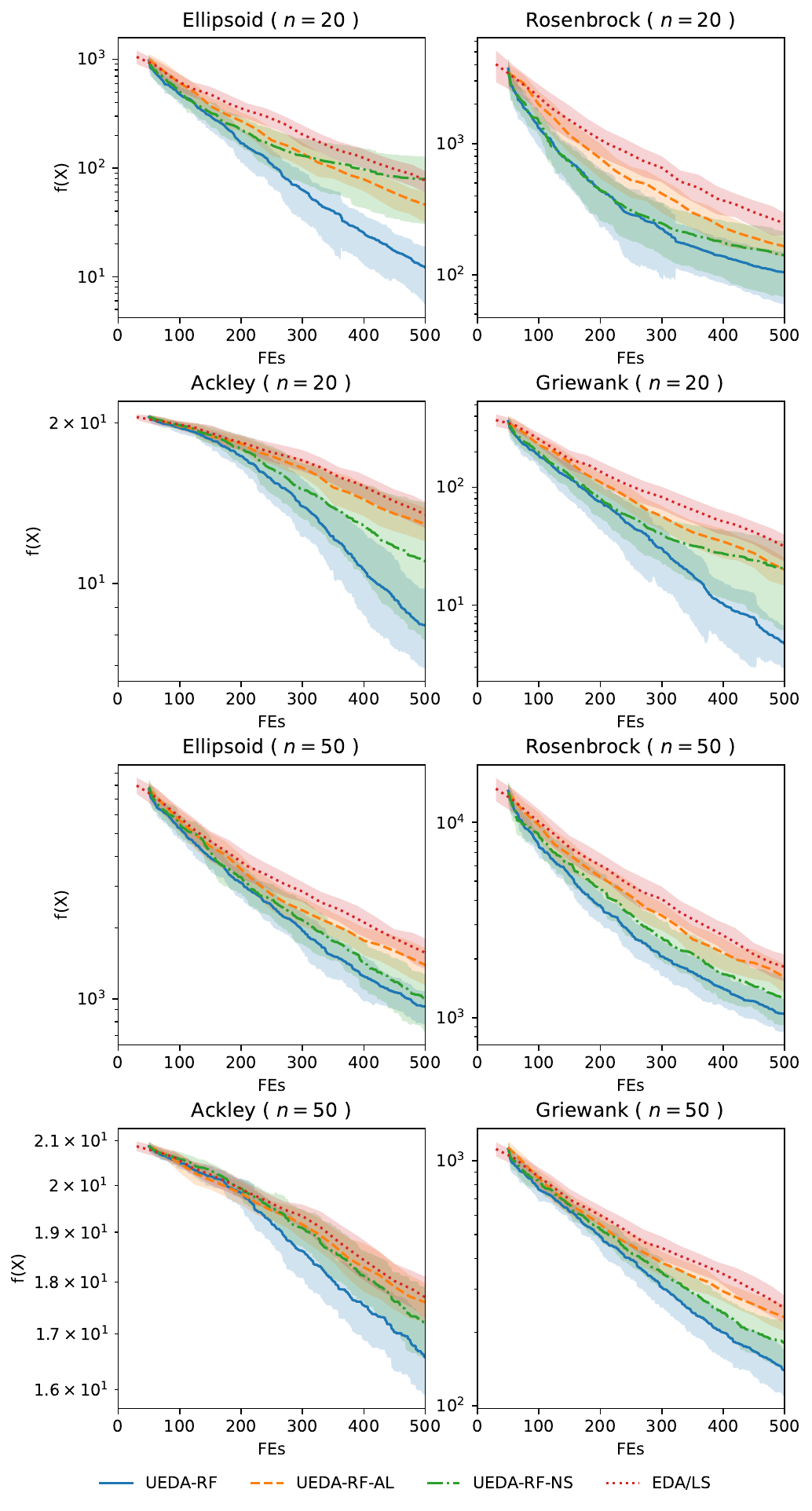}
    \caption{Comparison of EDA/LS and UEDAs on LZG test suits with $n=20,50$.}
    \label{fig:ablation}
  \end{figure}

The resulting curves of the objective values versus the number of fitness evaluations~(FEs) are illustrated in Fig.~\ref{fig:ablation}, where the lines represent the mean values of multiple trials, and the shaded areas indicate the standard deviation.

The experimental results are manifestly evident; UEDA-RF outperforms both variants as well as the baseline across all problems. This underscores the efficacy of UEDA and the significance of surrogate assisted selection and reproduction strategies. The performance of UEDA-RF-AL surpasses that of UEDA-RF-NS, and both excel over the baseline algorithm. Upon further analysis, it is apparent that EDA/LS was not tailored for expensive optimization problems, as a fixed number of individuals per generation is evaluated, incurring computational expenses. With limited computational resources (FEs = 500), the search is insufficient. UEDA-RF-ALL evaluates only half the number of individuals selected by the surrogate model per generation, effectively increasing the number of evolutionary generations and henceforth delivering superior performance. UEDA-RF-NS evaluates one individual per generation, which further economizes on computational resources; however, the loss of information from unevaluated solutions deteriorates the quality of solutions generated by reproduction, hence underperforming compared to the UEDA presented in this work. In summary, UEDA can effectively conserve computational expenses while ensuring the quality of new solutions, thereby enhancing search efficiency.

This section substantiates the effectiveness of the presented UEDA framework (utilizing RF as the surrogate model) through ablation studies. Variants are described as presented in Table~\ref{tab:ablation}. The ablation experiments investigate the advantages of UEDA in comparison to the baseline EDA/LS, as well as the impact on the algorithm's performance of evaluating $\mathbf{O}_{all}$ and the exclusion of unevaluated solutions in reproduction. The experiments are conducted on the LZG test suite, with problem dimensions set at $n=20$ and $n=50$. Each experimental set is independently replicated 30 times to mitigate the effects of randomness.

The resulting curves of the objective values versus the number of fitness evaluations (FEs) are illustrated in Fig.~\ref{fig:ablation}, where the lines represent the mean values of multiple trials, and the shaded areas indicate the standard deviation.

The experimental results are unequivocally clear; UEDA-RF outperforms both variants as well as the baseline across all problems. This underscores the efficacy of UEDA. The performance of UEDA-RF-AL surpasses that of UEDA-RF-NS, with both outperforming the baseline algorithm. Upon further analysis, it is evident that EDA/LS was not designed for expensive optimization problems, as it evaluates a fixed number of individuals per generation, leading to high computational costs. With limited computational resources (FEs = 500), the search proves to be insufficient. UEDA-RF-ALL evaluates only half the number of individuals selected by the surrogate model per generation, effectively increasing the number of evolutionary generations and thereby delivering superior performance. UEDA-RF-NS evaluates one individual per generation, which further economizes on computational resources; however, the loss of information from unevaluated solutions deteriorates the quality of the solutions generated by reproduction, thus underperforming compared to the UEDA framework presented in this work. In summary, UEDA can effectively conserve computational expenses while ensuring the quality of new solutions, thereby enhancing search efficiency.

\section{Conclusion}
\label{sec:conclusion}
This work began with an exploration of model uncertainty, comparing and contrasting the search behaviors of BO and SAEA. A visual analysis of one-dimensional functions has revealed that an evolutionary algorithm, which employs selection and reproduction based on populations, can mitigate the reliance on acquisition functions derived from model uncertainty. Furthermore, it can sidestep the misdirection caused by inherent model errors. Building on this insight, we proposed a novel evaluation-free iterative strategy that harnesses the strengths of population-based search while reducing evaluation costs. This strategy was implemented as the UEDA, which outperformed three mainstream BOs and two SAEAs in systematic experimental comparisons, thus demonstrating its effectiveness. Notably, UEDA also surpassed BOs in time efficiency by two orders of magnitude. Ablation studies further highlighted the significance of integrating evaluation-free solutions.

However, several avenues for future research remain open. First, the comparative analysis of BO and SAEA search behaviors was confined to one-dimensional functions, which may not capture the potential disparities in high-dimensional search spaces and complex functions. Therefore, a more systematic analysis is warranted. Second, as a general strategy, the application of evaluation-free solutions to other evolutionary search operators, such as genetic algorithms (GA) and differential evolution (DE), deserves further investigation. Lastly, the impact of different surrogate models on the performance of UEDA presents another promising area of study.

\bibliographystyle{IEEEtran}
\bibliography{mybibliography}

\end{document}